\documentclass{article}
\usepackage{spconf,amsmath,graphicx}
\usepackage{amsmath,amssymb}
\usepackage[hyphens]{url}
\usepackage{hyperref}
\usepackage{threeparttable}
\usepackage{multirow}
\usepackage{tikz}
\usepackage{pgfplots}
\pgfplotsset{compat=1.18}

\usepackage{subcaption}
\usepackage{threeparttable}
\hypersetup{colorlinks=true, breaklinks=true, linkcolor=blue, citecolor=blue, urlcolor=blue}
\usepackage[dvipsnames]{xcolor}
\newcommand{\cmark}{\textcolor{green!60!black}{\checkmark}}
\newcommand{\xmark}{\textcolor{red!70!black}{\texttimes}}

\title{Agreement-Driven Multi-View 3D Reconstruction for Live Cattle Weight Estimation}

\name{Rabin Dulal, Wenfeng Jia, Lihong Zheng, Jane Quinn}
\address{Charles Sturt University}
\begin{document}
%
\maketitle
\begin{abstract}
Accurate cattle live weight estimation is vital for livestock management, welfare, and productivity. Traditional methods, such as manual weighing using a walk-over weighing system or proximate measurements using body condition scoring, involve manual handling of stock and can impact productivity from both a stock and economic perspective. To address these issues, this study investigated a cost-effective, non-contact method for live weight calculation in cattle using 3D reconstruction. The proposed pipeline utilized multi-view RGB images with SAM 3D-based agreement-guided fusion, followed by ensemble regression. Our approach generates a single 3D point cloud per animal and compares classical ensemble models with deep learning models under low-data conditions. Results show that SAM 3D with multi-view agreement fusion outperforms other 3D generation methods, while classical ensemble models provide the most consistent performance for practical farm scenarios (R$^2$ = 0.69 $\pm$ 0.10, MAPE = 2.22 $\pm$ 0.56 \%), making this practical for on-farm implementation. These findings demonstrate that improving reconstruction quality is more critical than increasing model complexity for scalable deployment on farms where producing a large volume of 3D data is challenging.
\end{abstract}
\section{Introduction}
\label{sec:intro}

Live weight is a key metric of livestock performance in both beef and sheep production systems. As livestock systems face increasing sustainability challenges, accurate information on live weight gain is needed to guide management decisions, but preferably without impacting animal performance to achieve that measure. Currently, there is no reliable substitution for manual weighing of stock, as visual scoring systems can be imprecise, whilst also frequently requiring hands-on assessment that can impact live weight gain through handling.

Currently, live weight measurement of cattle is typically performed using direct methods, such as measurement using walk-over weighing scales placed in a crush or chute, or through indirect methods based on cattle morphology and physical presentation~\cite{wang2021asas}. Although direct weighing is accurate, manual body measurements and direct weighing are time-consuming, labour-intensive, costly, and require restraining cattle, which induces stress, reduces productivity, and poses safety risks to both producer and animal alike~\cite{ruchay2022live, ruchay2022comparative}. Consequently, recent studies~\cite{ruchay2022live, hou2023body, paudel2023deep, zheng2025weight, liu2024estimation, kwon2023deep, dang2023korean, okayama2021estimating, nguyen2023towards} have explored non-contact weight estimation using low-cost sensors and machine vision, enabling faster, safer, and less invasive weight prediction than scale-based methods.
 
\begin{figure}[t]
    \centering
    \includegraphics[width=1.0\linewidth]{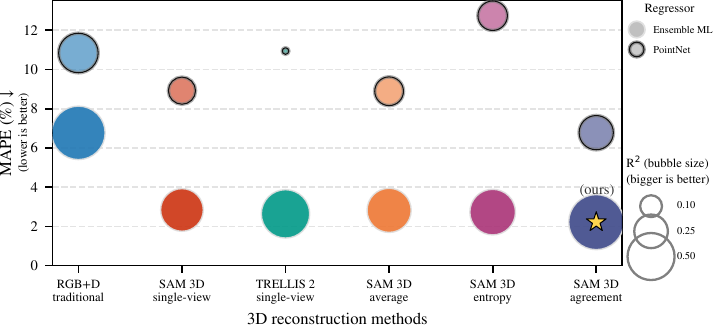}
    \vspace{-1em}
    \caption{MAPE and R$^2$ of weight estimation across 3D cattle reconstruction methods.}
    \label{fig:teaser}
    \vspace{-1.25em}
\end{figure}

In recent years, machine learning (ML) and deep learning (DL) approaches have been increasingly adopted for live weight prediction based on animal morphological characteristics. These methods offer a strong capability to model complex and non-linear relationships between body shape and weight from the data. 

3D data methodologies have gained increasing prominence for live weight estimation due to their ability to capture comprehensive spatial and volumetric information that cannot be reliably obtained from 2D images~\cite{paudel2023deep}. Unlike 2D vision systems, which are sensitive to viewpoint, lighting conditions, and occlusion, 3D sensing technologies provide accurate depth measurements and geometric representations of the animal’s body structure~\cite{zheng2025weight}. This enables more precise estimation of body volume, surface area, and shape-related features that are strongly correlated with live weight~\cite{hossain2025learning}. Consequently, 3D vision-based approaches offer improved robustness and generalization, particularly in unconstrained farm environments, making them well-suited for non-contact, automated, and accurate livestock weight prediction~\cite{hou2023body}. 

Recent studies on non-contact cattle weight estimation have used 3D data acquired from stereo camera systems, LiDAR sensors, and / or depth cameras such as Kinect and Intel RealSense to generate point clouds or 3D meshes of cattle body shape, size and condition~\cite{hou2023body,liu2024estimation,kwon2023deep,dang2023korean}. Weight prediction is typically performed using either classical ML models applied to handcrafted 3D features or DL methods that directly learn from 3D representations~\cite{paudel2023deep,zheng2025weight,okayama2021estimating,qi2017pointnet++}. Tab.~\ref{tab:existing_research} summarises existing non-contact live weight estimation studies.













\begin{table}[htbp]
\centering
\caption{Summary of non-contact animal live weight estimation studies. Our method achieves 3D reconstruction using only RGB images, with \emph{no need} for extra sensing hardware (Depth/LiDAR/Stereo).}
\vspace{-0.5em}
\label{tab:existing_research}
\scriptsize
\setlength{\tabcolsep}{3pt}
\renewcommand{\arraystretch}{1.1}
\resizebox{\columnwidth}{!}{%
\begin{tabular}{l c c l c c l}
\hline
Ref. & Data/animal & Animal & Weight Model & RGB & Depth info & Metrics \\
\hline
\cite{ruchay2022live}       & 11 & Cattle   & MRGBDM     & \cmark & \cmark & MAE, MAPE \\
\cite{hou2023body}          & 21 & Cattle   & PointNet++  & \xmark & \cmark & MAPE, RMSE \\
\cite{paudel2023deep}       & 5  & Pig      & PointNet    & \cmark & \cmark & $R^2$, RMSE \\
\cite{zheng2025weight}      & 5  & Chicken  & PointNet++  & \cmark & \cmark & MAE, MAPE \\
\cite{liu2024estimation}    & 50 & Pig      & MACNN       & \cmark & \cmark & MAE, MAPE, RMSE \\
\cite{kwon2023deep}         & 14 & Pig      & DNN         & \cmark & \cmark & $R^2$, MAE \\
\cite{dang2023korean}       & 4  & Cattle   & RF          & \xmark & \cmark & MAE, MAPE \\
\cite{okayama2021estimating}& 5  & Pig      & LR          & \xmark & \cmark & MAPE \\
\cite{nguyen2023towards}    & 4  & Pig      & ANN         & \cmark & \cmark & MAE, RMSE \\
\hline
\textbf{Ours}               & 1  & Cattle   & Ensemble ML  & \cmark & \xmark (\textbf{No need})      & $R^2$, MAPE, MAE \\
\hline
\end{tabular}%
}
\vspace{-1em}
\end{table}

The existing practical use of 3D-based live cattle weight estimation is limited by cattle movement, high hardware costs, complex installation and calibration, and the need for technical expertise, while DL methods further require large, well-labelled datasets that are difficult to obtain in real farm environments~\cite{dulal2025language, dulal2025ccomaml}. 

To address these challenges, this study makes three key contributions: \textbf{(i)} it proposes a cost-effective, non-contact pipeline that reconstructs accurate 3D representations from RGB images without requiring expensive sensors or preprocessing; \textbf{(ii)} it introduces an agreement-driven fusion strategy for SAM 3D-based multi-view 3D reconstruction to obtain a reliable and consistent 3D representation; and \textbf{(iii)} it develops an ensemble regression framework for cattle weight estimation that effectively leverages geometric features extracted from the reconstructed 3D models.




\section{Methodology}
\label{sec:methodology}
Building on recent advances in 3D reconstruction~\cite{chen2025sam,xiang2025trellis2,mv_sam3d_web}, we propose a non-contact pipeline that reconstructs accurate 3D representations of live cattle from RGB images and estimates their weight. 

\begin{figure*}[htbp]
    \centering
    \includegraphics[width=0.9\linewidth]{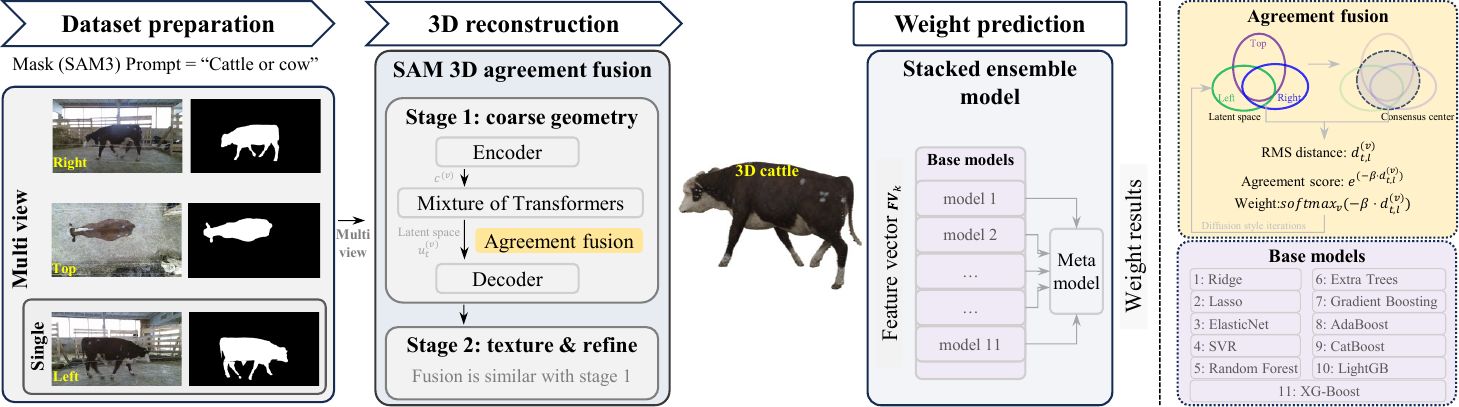}
    \vspace{-0.25em}
    \caption{Pipeline of agreement-driven multi-view 3D reconstruction for non-contact live cattle weight estimation.}
    \label{fig:pipeline}
    \vspace{-1em}
\end{figure*}

The overall pipeline is illustrated in Fig.~\ref{fig:pipeline}. \textbf{First}, the dataset~\cite{ruchay2020accurate} is collected using cameras, and masks are generated by SAM3~\cite{carion2025sam3} with a text prompt. \textbf{Second}, given the images and masks, 3D reconstruction is performed by SAM 3D with agreement fusion; we also reconstruct 3D live cattle using other methods, including conventional RGB+D~\cite{ruchay2020accurate} and TRELLIS2~\cite{xiang2025trellis2} etc., and compare these approaches through downstream live weight-estimation performance. \textbf{Finally}, cattle live weight is estimated using ensemble learning.

\subsection{Dataset Source Description}
The dataset used in this study is publicly available~\cite{ruchay2020accurate} and contains \textbf{103$\times$3} live cattle RGB images captured from left, right, and top views, along with corresponding 3D point clouds and live cattle weights. The point clouds include surrounding walls and floors, which were removed by us using RANSAC~\cite{cantzler1981random} to obtain clean representations.

\subsection{3D Data Generation}
\label{3Dgeneration}
SAM 3D~\cite{chen2025sam} is a two-stage generative foundation model for single-image 3D reconstruction, introduced in 2025. Its strong modeling capability enables the reconstruction of accurate 3D Gaussian point clouds from only RGB images, which in turn supports downstream tasks such as cattle weight estimation. However, SAM 3D originally supports only single-view image input. In contrast, \cite{mv_sam3d_web} extends SAM 3D to the multi-view setting by introducing simple average fusion and entropy-based fusion, thereby enabling multi-view 3D reconstruction based on SAM 3D.

Inspired by~\cite{mv_sam3d_web}, we introduce an agreement-weighted multi-view fusion strategy that operates within SAM~3D's iterative latent generation. The inputs are $V$ masked views
$\{(I^{(v)}, M^{(v)})\}_{v=1}^{V}$ (RGB + mask only). Each view is encoded into a conditioning representation:
\begin{equation}
\mathbf{c}^{(v)} = E\!\left(I^{(v)}, M^{(v)}\right), \qquad v=1,\dots,V,
\label{eq:cond_tokens}
\end{equation}
where $E(\cdot)$ denotes the SAM~3D's encoder and $\mathbf{c}^{(v)}$ are view-specific
conditioning tokens.

Let $\mathbf{x}_t$ denote the current latent state at step $t$. Conditioned on each view, the stage-specific transformers predict a view-dependent latent update:
\begin{equation}
\mathbf{u}^{(v)}_t = f_{\theta}\!\left(\mathbf{x}_t, t;\mathbf{c}^{(v)}\right), \qquad v=1,\dots,V.
\label{eq:view_updates}
\end{equation}

The agreement weights are computed and fused $\{\mathbf{u}^{(v)}_t\}$ before advancing the latent state.
Specifically, for each latent location $\ell$, we first form a multi-view consensus center (mean center) using the per-view update vectors: $\mathbf{u}^{(v)}_{t,\ell}\in\mathbb{R}^{D}$:
\begin{equation}
\mathbf{m}_{t,\ell}=\mathrm{Center}\!\left(\{\mathbf{u}^{(j)}_{t,\ell}\}_{j=1}^{V}\right).
\label{eq:consensus_center}
\end{equation}

Then, measuring how much each view deviates from this consensus via an RMS distance:
\begin{equation}
d^{(v)}_{t,\ell}=\sqrt{\tfrac{1}{D}\left\|\mathbf{u}^{(v)}_{t,\ell}-\mathbf{m}_{t,\ell}\right\|_2^2+\varepsilon},
\label{eq:deviation_score}
\end{equation}

where $D$ is the latent channel dimension and $\varepsilon$ is a small constant for numerical stability.
After that, converting deviations into an unnormalized agreement score:
\begin{equation}
agreement^{(v)}_{t,\ell}=\exp\!\left(-\beta\, d^{(v)}_{t,\ell}\right),
\label{eq:agreement_score}
\end{equation}

And apply a softmax over views to obtain fusion weights:
\begin{align}
w^{(v)}_{t,\ell}
&=\frac{agreement^{(v)}_{t,\ell}}{\sum_{j=1}^{V} agreement^{(j)}_{t,\ell}}
=\mathrm{softmax}_{v}\!\left(-\beta\, d^{(v)}_{t,\ell}\right),
\label{eq:agreement_weight}\\
\bar{\mathbf{u}}_{t,\ell}
&=\sum_{v=1}^{V} w^{(v)}_{t,\ell}\,\mathbf{u}^{(v)}_{t,\ell}.
\label{eq:agreement_fuse}
\end{align}

Here $\beta$ controls the sharpness of view selection.
The fused update $\bar{\mathbf{u}}_t$ is then used by the sampler to advance the latent state, producing Stage~1 and, analogously, the refined Stage~2 for decoding into 3D outputs.

\subsection{Weight Prediction}
In this study, live cattle weight is estimated from 3D point clouds using feature extraction, multiple regression models, and a stacked ensemble approach. Let the 3D point cloud of the $k$-th cattle be denoted by $\mathbf{P}_k$, defined as
\[
\mathbf{P}_k = \left\{ \mathbf{p}_{k,n} = (x_{k,n}, y_{k,n}, z_{k,n}) \in \mathbb{R}^3 \right\}_{n=1}^{N_k},
\]
where $k = 1, 2, \dots, K$ indexes the cattle instances, $n = 1, 2, \dots, N_k$ indexes the points within the $k$-th point cloud, and $N_k$ denotes the total number of points in $\mathbf{P}_k$.


For each cattle, a feature vector $\mathbf{FV}_k$ is computed to capture geometric, statistical, and density-based properties. These features include geometric descriptors (length, width, height, bounding box and convex hull volumes, and surface area), shape descriptors, spatial distribution (percentiles along the $x$, $y$, and $z$ axes), density and statistical moments (Z-axis densities in three vertical sections and mean and standard deviation along each axis). The feature selection is motivated by existing studies~\cite{ dang2023korean,okayama2021estimating,hossain2025learning,ruchay2020accurate}, which demonstrate the effectiveness of global body dimensions, shape descriptors, spatial distribution, and statistical moments for cattle weight estimation. These features are combined into a unified representation to capture both geometric structure and statistical characteristics.


\begin{equation}
\begin{aligned}
\mathbf{FV}_k \triangleq
\big[&
\mathbf{F}^{(g)}_k,\ \mathbf{F}^{(a)}_k,\ \mathbf{F}^{(q_x)}_k,\ \mathbf{F}^{(q_y)}_k,\ 
\mathbf{F}^{(q_z)}_k,\ \mathbf{F}^{(\rho)}_k,\ \mathbf{F}^{(\mu)}_k
\big],\\
\mathcal{P} &\triangleq \{10,25,50,75,90\},\\
\mathbf{F}^{(g)}_k &\triangleq [l,\ w,\ h,\ V_\mathrm{bbox},\ V_\mathrm{hull},\ A_\mathrm{surface}]^\top,\\
\mathbf{F}^{(a)}_k &\triangleq [\lambda_1/\lambda_2,\ \lambda_2/\lambda_3]^\top,\\
\mathbf{F}^{(q_x)}_k &\triangleq [x_p]_{p\in\mathcal{P}}^\top,\quad
\mathbf{F}^{(q_y)}_k \triangleq [y_p]_{p\in\mathcal{P}}^\top,\quad
\mathbf{F}^{(q_z)}_k \triangleq [z_p]_{p\in\mathcal{P}}^\top,\\
\mathbf{F}^{(\rho)}_k &\triangleq [\rho_\mathrm{Z1},\ \rho_\mathrm{Z2},\ \rho_\mathrm{Z3}]^\top,\\
\mathbf{F}^{(\mu)}_k &\triangleq [\mu_x,\ \sigma_x,\ \mu_y,\ \sigma_y,\ \mu_z,\ \sigma_z]^\top
\end{aligned}
\end{equation}

The resulting feature matrix $FV \in \mathbb{R}^{1 \times d}$ and target vector $T_{arget} \in \mathbb{R}$ (live weight in kg) are used to train a set of 11 base regression models,
\begin{equation}
\mathcal{G} = \left\{ g_m : \mathbb{R}^{1 \times d} \to \mathbb{R} \;\middle|\; m = 1,\dots,11 \right\}
\end{equation}
These models cover linear, kernel-based, and ensemble learning paradigms, enabling a robust evaluation of the relationship between 3D morphological features and cattle live weight. The regressors include Linear Regression, Ridge, Lasso, ElasticNet, SVR, Random Forest, Extra Trees, Gradient Boosting, AdaBoost, CatBoost, LightGBM, and XGBoost. For the $k$-th sample, the prediction of the $m$-th base model is
$\hat{T_{arget}}_{k,m} = f_m(\mathbf{FV}_k)$.

To leverage complementary model strengths, a stacked ensemble is constructed. Let the predictions of the top $m_{top}$ base models for the $k$-th sample be:
\begin{equation}
\hat{T_{arget}}_k^\text{base} =
\begin{bmatrix}
\hat{T_{arget}}_{k,1}, \dots, \hat{T_{arget}}_{k,m_{top}}
\end{bmatrix}^\top
\end{equation}
A Ridge regression model serves as the self-learner and produces the final prediction.
\begin{equation}
\hat{T_{arget}}_k^\text{final}
= g(\hat{T_{arget}}_k^\text{base})
= \mathbf{w}_\text{self}^\top \hat{T_{arget}}_k^\text{base} + b_\text{self}
\end{equation}
where $\mathbf{w}_\text{self}$ and $b_\text{self}$ are learned by minimizing the Ridge-regularized squared error,
\begin{equation}
\min_{\mathbf{w}_\text{self}}
\sum_{k=1}^{n} \big( {T_{arget}}_k - \hat{T_{arget}}_k^\text{final} \big)^2
+ \alpha \|\mathbf{w}_\text{self}\|_2^2
\end{equation}
with $\alpha$ fixed to 1.0 to provide stable $L2$ regularization under limited training data. The proposed methodology is evaluated using 5-fold cross-validation, and performance is assessed using the coefficient of determination (R$^2$), mean absolute error (MAE), and mean absolute percentage error (MAPE).

\section{Results and Discussion}
\label{sec:rd}
\subsection{3D Cattle Body Reconstruction}
\begin{figure}[htbp]
    \centering
    \includegraphics[width=1.0\linewidth]{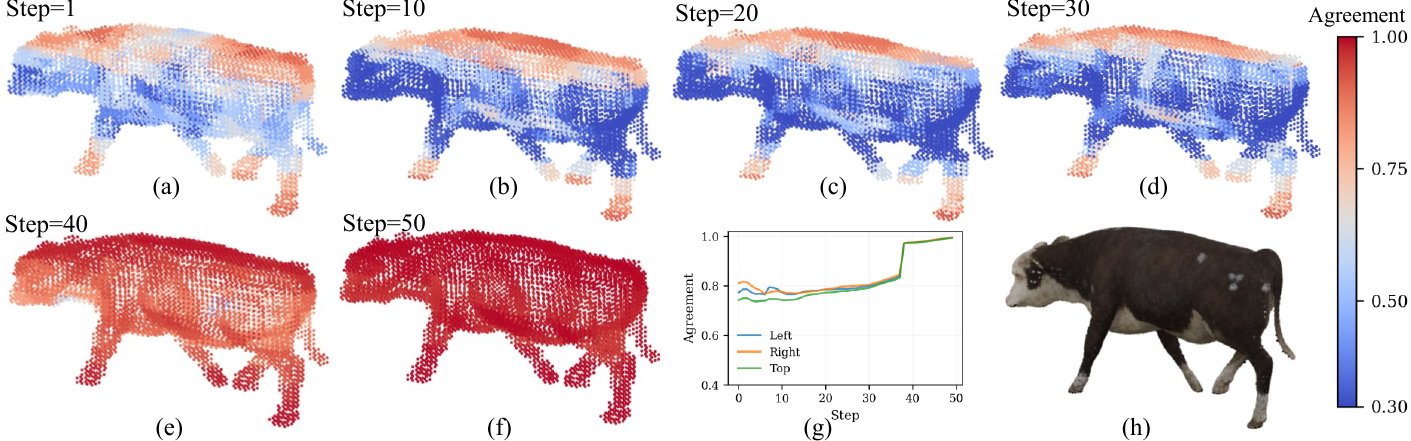}
    \vspace{-1.25em}
    \caption{Visualization of multi-view agreement fusion in SAM 3D coarse geometry stage latent space. \footnotesize{(a)--(f) shows the spatial distribution of the mean 3-view agreement score as the iterations progress. (g) is the per-view mean agreement score. (h) is the visualization of the final multi-view fused 3D Gaussian point cloud.}}
    \label{fig:multi_view}
    \vspace{-0.5em}
\end{figure}

As the agreement-based fusion proceeds (Sec.~\ref{3Dgeneration}), the reconstructed live cattle model becomes progressively more robust and complete. Fig.~\ref{fig:multi_view} visualizes the evolution of the agreement scores defined in Eq.~\ref{eq:agreement_score} across diffusion steps $t$ during the coarse Stage~1. Fig.~\ref{fig:multi_view} (a)-–(f) shows the spatial distribution of the mean multi-view $agreement_{t,\ell}$ for different cattle parts in the Stage~1 latent space as the iterations progress ($t$). Fig.~\ref{fig:multi_view}  (g) plots the per-view mean agreement score over diffusion steps. The agreement rises rapidly and approaches 1 at around $t\approx40$, indicating that $d^{(v)}_{t,\ell}$ in Eq.~\ref{eq:agreement_score} quickly decreases towards 0; consequently, all views strongly align with the consensus center $\mathbf{m}_{t,\ell}$ in latent space.

\begin{figure}[htbp]
    \centering
    \includegraphics[width=1.0\linewidth]{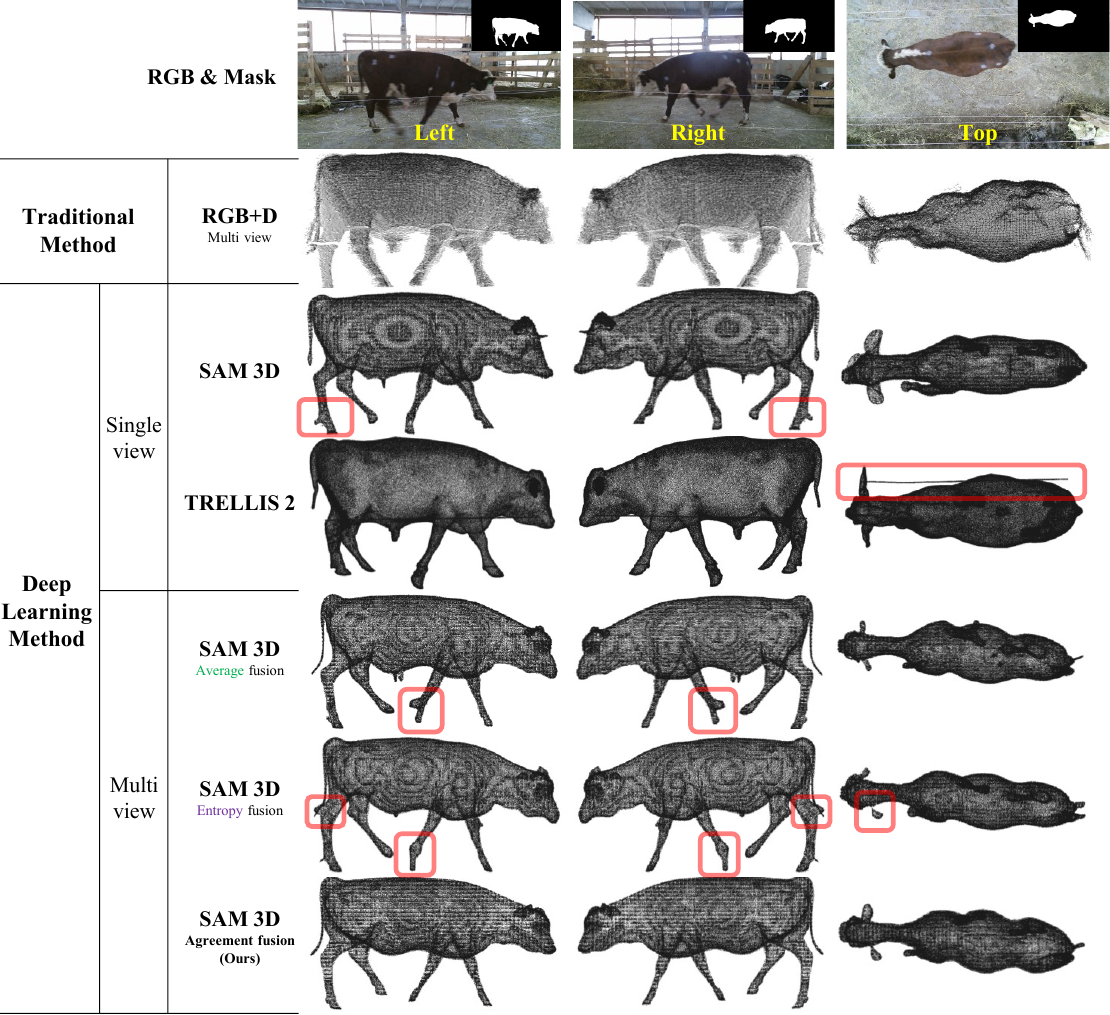}
    \vspace{-1em}
    \caption{Visual Comparison of different point clouds. Obvious structural anomalies are marked with red boxes. \footnotesize{SAM 3D outputs Gaussian point clouds; however, we ignore the Gaussian parameters and visualise them as uncoloured point clouds for comparison.}}
    \label{fig:3Ddatasets}
    \vspace{-1.25em}
\end{figure}
As presented in Fig.~\ref{fig:3Ddatasets}, it visually compares different methods used for 3D reconstruction. Among these methods, RGB+D reconstruction yields the most natural cattle shape, while details are coarse, edges are blurry, and occlusions noticeably degrade the results. Among DL methods, TRELLIS2 achieves the best visual quality but is highly sensitive to occlusions. Moreover, for SAM 3D-based approaches, multi-view agreement fusion performs best overall, producing more consistent cattle shapes with fewer artifacts.


\subsection{Cattle Weight Regression}
\begin{figure}[htbp]
    \centering
    \vspace{-6pt}
    \includegraphics[width=0.9\linewidth]{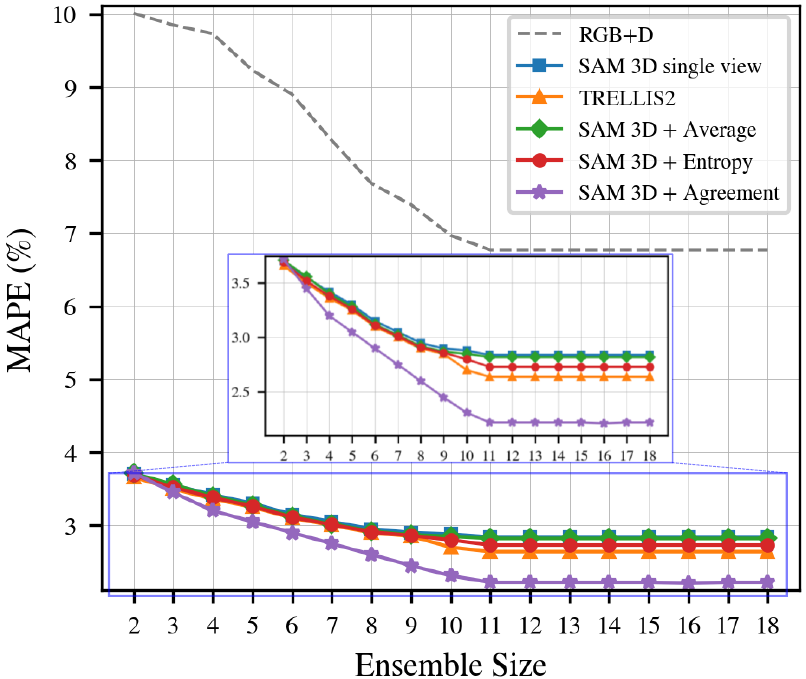}
    \vspace{-6pt}
    \caption{Effect of ensemble size on MAPE (\%) for different 3D dataset. 
    \footnotesize{Performance improves with increasing top ML models but stabilizes after Top-11 for both RGB+D and DL-based 3D methods.}}
    \label{fig:MAPE_plot}
    \vspace{-6pt}
\end{figure}

18 ML models and 2 DL models were evaluated. The ensemble of the top 11 ML models achieved the best performance, as including additional models beyond the top 11 did not yield significant improvement, as shown in Fig~\ref{fig:MAPE_plot}. Notably, weight estimates from DL-based 3D reconstructions significantly outperform those obtained using conventional RGB+D methods. Detailed results are provided in the Supplementary Material.

Tab.~\ref{tab:regression_core} presents the results of cattle weight prediction using the ensemble model, PointNet and PointNet++. Classical ensemble models consistently outperform DL models (PointNet and PointNet++) on different cattle point cloud datasets. The ensemble model trained on \textbf{SAM 3D with agreement fusion} dataset achieves the highest R$^2$ of 0.69$\pm$0.10, the best MAE of 9.16$\pm$2.32 kg, and MAPE of 2.22$\pm$0.56\%. These results demonstrate that SAM 3D-based multi-view reconstruction with agreement fusion produces the most accurate and reliable 3D representations, while classical ensemble models outperform DL approaches in low-data scenarios.

\begin{figure}[htbp]
    \centering
    \includegraphics[width=0.9\linewidth]{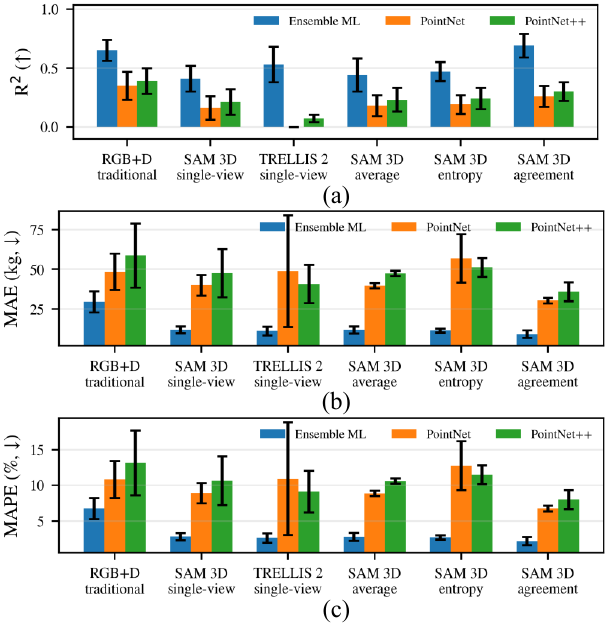}
    \vspace{-0.5em}
    \caption{Comparison of weight estimation performance across models and 3D reconstruction methods. \footnotesize{(a) results of R$^2$; (b) results of MAE; (c) results of MAPE.}}
    \label{fig:performances}
    \vspace{-1.5em}
\end{figure}

An ensemble of classical ML models consistently outperforms DL models across all 3D reconstruction methods, as shown in Tab.~\ref{tab:regression_core} and Fig.~\ref{fig:performances}. On conventional RGB+D datasets, the ensemble regressor attains an R$^2$ of 0.65$\pm$0.09, outperforming PointNet (0.35$\pm$0.12) and PointNet++. 
In the other datasets, in terms of MAE and MAPE, it also shows that ensemble ML clearly outperforms the DL models. This underperformance stems primarily from the scarcity of training data: each cattle has only one point cloud, which is insufficient for deep networks to learn meaningful 3D spatial patterns, leading to overfitting and poor generalization.

\begin{table}[htbp]
\centering
\caption{Regression results on point clouds using 5-fold cross-validation (mean$\pm$std). This table reports only Ensemble ML, PointNet, PointNet++. The best 3D reconstruction method is marked in \textbf{\textcolor{red}{red}}, the second best is marked in \textcolor{blue}{blue}; \footnotesize{full results are provided in the supplementary material}.}
\label{tab:regression_core}
\begin{threeparttable}
\tiny
\setlength{\tabcolsep}{2pt}
\renewcommand{\arraystretch}{1.0}
\resizebox{0.95\columnwidth}{!}{%
\begin{tabular}{l|ccc}
\hline
Method & Ensemble ML & PointNet & PointNet++ \\
\hline

\multicolumn{4}{c}{R$^2$ ($\uparrow$)} \\
\hline
RGB+D
& \textcolor{blue}{0.65$\pm$0.09}   & 0.35$\pm$0.12 & 0.39$\pm$0.11 \\
SAM 3D single view
& 0.41$\pm$0.11  & 0.16$\pm$0.10 & 0.21$\pm$0.11 \\
TRELLIS2
& 0.53$\pm$0.15  & -0.41$\pm$0.18 & 0.07$\pm$0.03 \\
SAM 3D + average
& 0.44$\pm$0.14  & 0.18$\pm$0.09 & 0.23$\pm$0.10 \\
SAM 3D + entropy
& 0.47$\pm$0.08  & 0.19$\pm$0.08 & 0.24$\pm$0.09 \\
\textbf{SAM 3D + agreement}
& \textbf{\textcolor{red}{0.69$\pm$0.10}} & 0.26$\pm$0.09  & 0.30$\pm$0.08  \\
\hline

\multicolumn{4}{c}{MAE (kg, $\downarrow$)} \\
\hline
RGB+D
& 29.51$\pm$6.67 & 48.5$\pm$11.5 & 58.8$\pm$20.2 \\
SAM 3D single view
& 11.83$\pm$2.04 & 39.9$\pm$6.5 & 47.6$\pm$15.2 \\
TRELLIS2
& \textcolor{blue}{11.12$\pm$2.68} & 48.9$\pm$35.3 & 40.8$\pm$12.0 \\
SAM 3D + average
& 11.77$\pm$2.21 & 39.7$\pm$1.7 & 47.5$\pm$1.6 \\
SAM 3D + entropy
& 11.38$\pm$1.21 & 56.9$\pm$15.2 & 51.3$\pm$5.9 \\
\textbf{SAM 3D + agreement}
& \textbf{\textcolor{red}{9.16$\pm$2.32}} & 30.3$\pm$1.7 & 35.9$\pm$6.0 \\
\hline

\multicolumn{4}{c}{MAPE (\%, $\downarrow$)} \\
\hline
RGB+D
& 6.77$\pm$1.46 & 10.84$\pm$2.59 & 13.16$\pm$4.52 \\
SAM 3D single view
& 2.84$\pm$0.49 & 8.92$\pm$1.45 & 10.66$\pm$3.41 \\
TRELLIS2
& \textcolor{blue}{2.64$\pm$0.64} & 10.94$\pm$7.89 & 9.12$\pm$2.90 \\
SAM 3D + average
& 2.82$\pm$0.53 & 8.89$\pm$0.38 & 10.62$\pm$0.35 \\
SAM 3D + entropy
& 2.73$\pm$0.29 & 12.74$\pm$3.40 & 11.48$\pm$1.31 \\
\textbf{SAM 3D + agreement}
& \textbf{\textcolor{red}{2.22$\pm$0.56}} & 6.78$\pm$0.38 & 8.02$\pm$1.34 \\
\hline
\end{tabular}%
}
\end{threeparttable}
\end{table}

As illustrated in Tab.~\ref{tab:regression_core} and Fig.~\ref{fig:teaser}, DL-based cattle 3D reconstruction methods substantially outperform conventional RGB+D approache. Across the considered 3D reconstruction methods, SAM 3D-based multi-view agreement fusion provides the most informative 3D representation for weight prediction, consistently improving MAE, MAPE and R$^2$. Notably, in terms of R$^2$, agreement fusion (0.69$\pm$0.10) is the only method that surpasses the conventional RGB+D baseline (0.65$\pm$0.09). 
The remaining DL-based 3D methods achieve lower R$^2$ values. This suggests that agreement fusion provides not only higher reconstruction quality but also stronger stability for downstream weight estimation.

In terms of MAE and MAPE, Tab.~\ref{tab:regression_core} and Fig.~\ref{fig:performances}(b, c) show that SAM 3D with multi-view agreement fusion dataset provides the best performance. Among the SAM 3D-based multi-view reconstruction variants, only agreement fusion achieves a clear improvement in weight estimation over single-view SAM 3D. However, the gain from the conventional RGB+D baseline (MAE=29.51$\pm$6.67, MAPE=6.77$\pm$1.46) to SAM 3D is much larger than the additional gain from single-view SAM 3D to multi-view agreement fusion (MAE=9.16$\pm$2.32, MAPE=2.22$\pm$0.56). Even so, the higher R$^2$ achieved by the agreement fusion method (R$^2$=0.69$\pm$0.10) suggests markedly greater reliability for practical cattle weight estimation in real applications.


\section{Conclusion}
\label{sec:conclusion}
This study presented a cost-effective, non-contact pipeline for live weight estimation from RGB images. Specifically, we developed a multi-view 3D reconstruction model that extends SAM 3D via a novel agreement fusion strategy. Notably, our results demonstrate that agreement fusion robustly utilizes cross-view information, leading to significant gains in reconstruction accuracy and reliability. Furthermore, the results confirmed that DL-based 3D reconstruction methods produce superior representation for live weight estimation compared to conventional RGB+D methods. Regarding the live weight estimation, the ensemble learning model, combining 11 ML models, was found to estimate live weights more accurately than DL model in low-data scenarios.

Overall, the proposed pipeline significantly reduces the time, cost, manpower, and effort required to generate 3D models. Consequently, it offers a robust solution for accurate individual animal live weight estimation, which is highly suitable for practical deployment in real world livestock management situations.

\newpage
\textbf{SUPPLEMENTARY MATERIAL}

\section{Selection Criteria of ML models}
For effective weight prediction, 18 regression models were used in this study. These include Extra Trees~\cite{geurts2006extremely}, Random Forest~\cite{breiman2001random}, AdaBoost~\cite{solomatine2004adaboost}, Gradient Boosting~\cite{friedman2001greedy}, XGBoost~\cite{chen2016xgboost}, CatBoost~\cite{hancock2020catboost}, LightGBM~\cite{ke2017lightgbm}, Linear Regression~\cite{weisberg2005applied}, Ridge~\cite{hoerl1970ridge}, Lasso~\cite{tibshirani1996regression}, Elastic Net~\cite{hans2011elastic}, Support Vector Regression (SVR)~\cite{smola2004tutorial}, Nu-Support Vector Regression (NuSVR)~\cite{hao2016pair}, Decision Tree~\cite{maimon2014data}, k-Nearest Neighbors (KNN)~\cite{fix1985discriminatory}, Huber Regression~\cite{huber1992robust}, RANSAC~\cite{cantzler1981random}, and Ridge Regression~\cite{hilt1977ridge}.

All regression models were trained using the original point cloud data (RGB+D) to identify and rank them from best to worst, as summarized in Table~\ref{regression_hierarchy}. To evaluate the benefits of ensembling, results for all models are also presented in Table~\ref{ensemble_all_models}.

\begin{table}[htbp]
\centering
\caption{Performance comparison of regression models on original point cloud data (RGB+D), ranked by MAPE (\%).}
\label{regression_hierarchy}
\scriptsize
\setlength{\tabcolsep}{4pt}
\renewcommand{\arraystretch}{1.1}
\begin{tabular}{clccc}
\hline
Rank & Model & $R^2$ & MAE (kg) & MAPE (\%) \\
\hline
1  & ExtraTrees        & 0.3044  & 47.20  & 10.57 \\
2  & RandomForest      & 0.2585  & 48.20  & 10.82 \\
3  & AdaBoost          & 0.2603  & 50.01  & 11.21 \\
4  & XGBoost           & 0.1756  & 50.60  & 11.31 \\
5  & GradientBoosting  & 0.2091  & 50.70  & 11.32 \\
6  & CatBoost          & 0.1800  & 52.10  & 11.64 \\
7  & Ridge             & 0.2028  & 53.60  & 12.00 \\
8  & LightGBM          & 0.0100  & 54.90  & 12.27 \\
9  & Lasso             & 0.1108  & 55.50  & 12.41 \\
10 & ElasticNet        & 0.0781  & 56.50  & 12.63 \\
11 & SVR               & -0.0442 & 60.01  & 13.40 \\
12 & Linear Regression & -0.2089 & 61.03  & 13.89 \\
13 & NuSVR             & -0.0534 & 64.61  & 14.51 \\
14 & DecisionTree      & -0.5813 & 65.12  & 14.62 \\
15 & Huber             & 0.1070  & 67.40  & 15.00 \\
16 & KNN               & -0.1140 & 69.48  & 15.25 \\
17 & BayesianRidge     & -0.1136 & 69.70  & 15.47 \\
18 & RANSAC            & -7.1467 & 134.91 & 31.81 \\
\hline
\end{tabular}
\end{table}

The results of the ensemble of top 2 to top 18 models are as presented in Table~\ref{ensemble_all_models}.
\begin{table}[htbp]
\centering
\caption{Performance of ensemble models formed by progressively adding top-ranked regressors on RGB+D data.}
\label{ensemble_all_models}
\scriptsize
\setlength{\tabcolsep}{5pt}
\renewcommand{\arraystretch}{1.1}
\begin{tabular}{lccc}
\hline
Ensemble Size & $R^2$ & MAE (kg) & MAPE (\%) \\
\hline
Top2  & 0.3200 & 46.70 & 10.01 \\
Top3  & 0.3783 & 41.32 & 9.85 \\
Top4  & 0.3862 & 40.75 & 9.73 \\
Top5  & 0.5300 & 40.42 & 9.23 \\
Top6  & 0.5500 & 37.88 & 8.90 \\
Top7  & 0.5510 & 35.69 & 8.27 \\
Top8  & 0.5928 & 33.18 & 7.68 \\
Top9  & 0.6100 & 32.07 & 7.39 \\
Top10 & 0.6500 & 30.35 & 6.97 \\
Top11 & 0.6500 & 29.51 & 6.77 \\
Top12 & 0.6481 & 29.48 & 6.77 \\
Top13 & 0.6571 & 30.01 & 6.77 \\
Top14 & 0.6439 & 29.91 & 6.77 \\
Top15 & 0.6550 & 29.16 & 6.77 \\
Top16 & 0.6491 & 29.06 & 6.77 \\
Top17 & 0.6473 & 29.96 & 6.77 \\
Top18 & 0.6498 & 29.43 & 6.77 \\
\hline
\end{tabular}
\end{table}

The results show that after ensembling top11 models, the performance does not gain significantly, and there is no value added in the performance by adding more than 11 models. 
\section{Cattle Weight Regression}
The results of cattle weight prediction are presented in Table~\ref{tab:regression_results}. The table shows that the ensemble of the 11 top-performing regression models consistently outperforms the individual models. Moreover, the 3D point cloud generated using SAM-based multi-view reconstruction with agreement fusion achieves the best reconstruction quality, which is reflected in the most accurate weight prediction results.

\begin{table*}[htbp]
\centering
\caption{Performance comparison of regression models on point cloud data using 5-fold cross-validation (mean $\pm$ SD).}
\label{tab:regression_results}
\scriptsize
\setlength{\tabcolsep}{1pt}
\resizebox{\textwidth}{!}{%
\begin{tabular}{l|cccccccccccccc}
\hline
Metric / Model &
ExtraTrees & RF & AdaBoost & GB & Ridge & CatBoost & XGBoost & Lasso & ElasticNet & LightGBM & SVR & PointNet & PointNet++ & Ensemble \\
\hline

\multicolumn{15}{c}{Original point cloud} \\ \hline
R$^2$ &
0.30$\pm$0.13 & 0.26$\pm$0.11 & 0.26$\pm$0.11 & 0.21$\pm$0.15 &
0.20$\pm$0.12 & 0.18$\pm$0.14 & 0.18$\pm$0.17 &
0.11$\pm$0.13 & 0.08$\pm$0.10 & 0.01$\pm$0.18 &
-0.04$\pm$0.03 & 0.35$\pm$0.12 & 0.39$\pm$0.11 & 0.65$\pm$0.09 \\

MAE (kg) &
47.20$\pm$13.80 & 48.20$\pm$14.00 & 50.00$\pm$15.20 & 50.70$\pm$14.10 &
53.60$\pm$20.30 & 52.10$\pm$16.00 & 50.60$\pm$14.10 &
55.50$\pm$17.10 & 56.50$\pm$16.10 & 54.90$\pm$13.50 &
60.10$\pm$19.70 & 48.50$\pm$11.50 & 58.80$\pm$20.20 & 29.51$\pm$6.67 \\

MAPE (\%) &
10.57$\pm$3.18 & 10.82$\pm$3.26 & 11.21$\pm$3.52 & 11.32$\pm$3.14 &
12.00$\pm$4.54 & 11.64$\pm$3.57 & 11.31$\pm$3.15 &
12.41$\pm$4.63 & 12.63$\pm$4.28 & 12.27$\pm$3.02 &
13.40$\pm$4.39 & 10.84$\pm$2.59 & 13.16$\pm$4.52 & 6.77$\pm$1.46 \\

\hline
\multicolumn{15}{c}{\textbf{SAM multiview + agreement fusion}} \\ \hline
R$^2$ &
0.25$\pm$0.25 & 0.22$\pm$0.24 & 0.21$\pm$0.27 & 0.49$\pm$0.45 &
0.02$\pm$0.15 & 0.41$\pm$0.44 & 0.32$\pm$0.24 &
0.02$\pm$0.09 & 0.02$\pm$0.05 & -0.46$\pm$0.37 &
-0.08$\pm$0.09 & 0.26$\pm$0.09 & 0.30$\pm$0.08 & 0.69$\pm$0.10 \\

MAE (kg) &
18.00$\pm$2.20 & 17.90$\pm$1.50 & 18.60$\pm$1.80 & 20.80$\pm$2.80 &
16.80$\pm$2.10 & 19.90$\pm$2.40 & 19.50$\pm$2.40 &
17.20$\pm$1.80 & 17.50$\pm$1.80 & 20.50$\pm$1.80 &
16.50$\pm$1.60 & 30.30$\pm$1.70 & 35.90$\pm$6.00 & 9.16$\pm$2.32 \\

MAPE (\%) &
4.25$\pm$0.52 & 4.20$\pm$0.36 & 4.17$\pm$0.34 & 4.66$\pm$0.61 &
3.76$\pm$0.47 & 4.45$\pm$0.50 & 4.44$\pm$0.56 &
3.83$\pm$0.42 & 3.90$\pm$0.41 & 4.58$\pm$0.36 &
3.73$\pm$0.36 & 6.78$\pm$0.38 & 8.02$\pm$1.34 & 2.22$\pm$0.56 \\

\hline
\multicolumn{15}{c}{SAM Single View} \\ \hline
R$^2$ &
0.12$\pm$0.21 & 0.09$\pm$0.14 & 0.07$\pm$0.16 & 0.04$\pm$0.14 &
0.10$\pm$0.09 & 0.17$\pm$0.16 & 0.20$\pm$0.16 &
0.05$\pm$0.05 & 0.04$\pm$0.04 & -0.08$\pm$0.09 &
-0.08$\pm$0.09 & 0.16$\pm$0.10 & 0.21$\pm$0.11 & 0.41$\pm$0.11 \\

MAE (kg) &
18.60$\pm$3.60 & 18.30$\pm$2.80 & 17.90$\pm$2.70 & 20.70$\pm$2.80 &
17.60$\pm$2.40 & 19.10$\pm$2.90 & 20.20$\pm$2.10 &
18.30$\pm$2.00 & 18.30$\pm$1.90 & 19.00$\pm$2.80 &
16.80$\pm$1.60 & 39.90$\pm$6.50 & 47.60$\pm$15.20 & 11.83$\pm$2.04 \\

MAPE (\%) &
4.16$\pm$0.81 & 4.06$\pm$0.66 & 4.00$\pm$0.63 & 4.63$\pm$0.63 &
3.96$\pm$0.55 & 4.27$\pm$0.67 & 4.51$\pm$0.43 &
3.95$\pm$0.47 & 3.95$\pm$0.44 & 4.26$\pm$0.65 &
3.74$\pm$0.36 & 8.92$\pm$1.45 & 10.66$\pm$3.41 & 2.84$\pm$0.49 \\

\hline
\multicolumn{15}{c}{SAM + average fusion} \\ \hline
R$^2$ &
0.22$\pm$0.22 & 0.28$\pm$0.26 & 0.26$\pm$0.23 & 0.27$\pm$0.16 &
0.06$\pm$0.05 & 0.22$\pm$0.11 & 0.26$\pm$0.13 &
0.04$\pm$0.04 & 0.04$\pm$0.04 & -0.58$\pm$0.50 &
-0.09$\pm$0.09 & 0.18$\pm$0.09 & 0.23$\pm$0.10 & 0.44$\pm$0.14 \\

MAE (kg) &
19.10$\pm$2.30 & 19.30$\pm$1.90 & 19.10$\pm$2.30 & 22.80$\pm$2.10 &
17.60$\pm$1.70 & 19.00$\pm$2.00 & 22.80$\pm$3.40 &
18.60$\pm$1.80 & 18.60$\pm$1.80 & 21.30$\pm$3.30 &
16.70$\pm$1.50 & 39.70$\pm$1.70 & 47.50$\pm$1.60 & 11.77$\pm$2.21 \\

MAPE (\%) &
4.26$\pm$0.51 & 4.32$\pm$0.42 & 4.30$\pm$0.50 & 5.08$\pm$0.48 &
3.95$\pm$0.41 & 4.24$\pm$0.45 & 4.93$\pm$0.70 &
3.98$\pm$0.41 & 3.97$\pm$0.41 & 4.77$\pm$0.72 &
3.74$\pm$0.35 & 8.89$\pm$0.38 & 10.62$\pm$0.35 & 2.82$\pm$0.53 \\

\hline
\multicolumn{15}{c}{SAM + entropy fusion} \\ \hline
R$^2$ &
0.09$\pm$0.18 & 0.12$\pm$0.12 & 0.22$\pm$0.12 & 0.33$\pm$0.21 &
0.01$\pm$0.08 & 0.11$\pm$0.11 & 0.15$\pm$0.09 &
0.01$\pm$0.05 & 0.02$\pm$0.04 & -0.17$\pm$0.21 &
-0.08$\pm$0.09 & 0.19$\pm$0.08 & 0.24$\pm$0.09 & 0.47$\pm$0.08 \\

MAE (kg) &
17.50$\pm$0.70 & 17.60$\pm$0.90 & 18.20$\pm$1.30 & 19.40$\pm$0.80 &
17.50$\pm$2.60 & 18.10$\pm$1.60 & 19.70$\pm$1.00 &
17.50$\pm$1.60 & 17.60$\pm$1.60 & 18.50$\pm$1.30 &
16.50$\pm$1.50 & 56.90$\pm$15.20 & 51.30$\pm$5.90 & 11.38$\pm$1.21 \\

MAPE (\%) &
3.87$\pm$0.15 & 3.98$\pm$0.24 & 4.16$\pm$0.33 & 4.31$\pm$0.22 &
3.86$\pm$0.58 & 4.04$\pm$0.39 & 4.66$\pm$0.17 &
3.86$\pm$0.41 & 3.90$\pm$0.40 & 4.18$\pm$0.34 &
3.72$\pm$0.35 & 12.74$\pm$0.34 & 11.48$\pm$0.31 & 2.73$\pm$0.29 \\

\hline
\multicolumn{15}{c}{TRELLIS2} \\ \hline
R$^2$ &
0.07$\pm$0.12 & 0.11$\pm$0.13 & 0.08$\pm$0.12 & 0.02$\pm$0.15 &
-0.01$\pm$0.08 & 0.10$\pm$0.13 & -0.53$\pm$0.39 &
-0.01$\pm$0.24 & -0.02$\pm$0.04 & -0.17$\pm$0.21 &
-0.09$\pm$0.08 & -0.41$\pm$0.18 & 0.07$\pm$0.03 & 0.53$\pm$0.15 \\

MAE (kg) &
16.30$\pm$3.00 & 16.50$\pm$2.80 & 16.70$\pm$2.60 & 16.40$\pm$2.50 &
16.70$\pm$2.40 & 16.60$\pm$2.80 & 16.30$\pm$2.40 &
32.30$\pm$28.20 & 30.50$\pm$24.20 & 17.00$\pm$0.80 &
16.60$\pm$1.50 & 48.90$\pm$35.30 & 40.80$\pm$12.00 & 11.12$\pm$2.68 \\

MAPE (\%) &
3.68$\pm$0.68 & 3.74$\pm$0.65 & 3.87$\pm$0.63 & 3.78$\pm$0.60 &
3.86$\pm$0.58 & 3.76$\pm$0.65 & 3.79$\pm$0.57 &
7.42$\pm$6.81 & 6.93$\pm$5.78 & 4.13$\pm$0.18 &
3.74$\pm$0.35 & 10.94$\pm$7.89 & 9.12$\pm$2.90 & 2.64$\pm$0.64 \\

\hline
\end{tabular}%
}
\end{table*}
\bibliographystyle{IEEEbib}
\bibliography{strings,refs}

\end{document}